\documentclass[runningheads]{llncs}
\usepackage{graphicx}
\usepackage{soul}
\usepackage{xcolor}
\usepackage{amsmath}
\usepackage{amsfonts}
\usepackage{wrapfig}
\usepackage{bm}
\usepackage[framemethod=TikZ]{mdframed}

\usepackage{graphicx}
\usepackage{enumitem}
\usepackage[utf8]{inputenc} 
\usepackage[T1]{fontenc}    
\usepackage{hyperref}       
\usepackage{url}            
\usepackage{booktabs}       
\usepackage{nicefrac}       
\usepackage{microtype}      

\begin{document}
\title{Ruffle\&Riley: Insights from Designing and Evaluating a Large Language Model-Based Conversational Tutoring System}
%
\titlerunning{Designing and Evaluating a LLM-based CTS}
%

\author{Robin Schmucker\inst{1} \and
Meng Xia\inst{2} \and
Amos Azaria\inst{3} \and
Tom Mitchell\inst{1}}
\authorrunning{Schmucker et al.}
%
\institute{Carnegie Mellon University, Pittsburgh, PA 15213, USA \and
Texas A\&M University, College Station, TX 77843, USA \and
Ariel University, Ariel 4070000, Israel\\
\email{\{rschmuck, mitchell\}@cs.cmu.edu}}
%

%
\maketitle              
\begin{abstract}
Conversational tutoring systems (CTSs) offer learning experiences through interactions based on natural language. They are recognized for promoting cognitive engagement and improving learning outcomes, especially in reasoning tasks. Nonetheless, the cost associated with authoring CTS content is a major obstacle to widespread adoption and to research on effective instructional design. In this paper, we discuss and evaluate a novel type of CTS that leverages recent advances in large language models (LLMs) in two ways: First, the system enables AI-assisted content authoring by inducing an easily editable tutoring script automatically from a lesson text. Second, the system automates the script orchestration in a learning-by-teaching format via two LLM-based agents (Ruffle\&Riley) acting as a student and a professor. The system allows for free-form conversations that follow the ITS-typical inner and outer loop structure. We evaluate Ruffle\&Riley's ability to support biology lessons in two between-subject online user studies (N = 200) comparing the system to simpler QA chatbots and reading activity. Analyzing system usage patterns, pre/post-test scores and user experience surveys, we find that Ruffle\&Riley users report high levels of engagement, understanding and perceive the offered support as helpful. Even though Ruffle\&Riley users require more time to complete the activity, we did not find significant differences in short-term learning gains over the reading activity. Our system architecture and user study provide various insights for designers of future CTSs. We further open-source our system to support ongoing research on effective instructional design of LLM-based learning technologies.
 \and large language models
\end{abstract}


\section{Introduction}
\label{sec:introduction}

Intelligent tutoring systems (ITSs) are an transformative educational technology that provides millions of learners worldwide with access to learning materials and affordable adaptive instruction. ITSs can, in certain contexts, be as effective as human tutors~\cite{Kulik2016:Effectiveness} and can take on an important role in mitigating the educational achievement gap~\cite{Huang2016:Intelligent}. However, despite their potential, one major obstacle to the widespread adoption of ITS technologies, is the large costs associated with content development. Depending on the depth of instructional design and available authoring tools, preparing one hour of ITS content can take designers hundreds of hours~\cite{Aleven2016:Example}. This significant investment often necessitates that ITSs focus on core subject areas and cater to larger demographic groups, limiting the breadth of topics covered and the diversity of learners adequately served.

Conversational tutoring systems (CTSs) are a type of ITS that engages with learners in natural language. Various studies have confirmed the benefits of CTSs, across multiple domains, particularly on learning outcomes in reasoning tasks~\cite{Paladines2020:Systematic}. Still, many existing CTSs struggle to maintain coherent free-form conversations and understand the learners' responses due to limitations imposed by their underlying natural language processing (NLP) techniques~\cite{Nye2014:Autotutor}. In this paper, we introduce and evaluate a new type of CTS that draws inspiration from design principles of earlier CTSs~\cite{Nye2014:Autotutor,leelawong2008designing} while leveraging recent advances in large language models (LLMs) to accelerate content authoring and to facilitate free-form conversational tutoring. Our main contributions include:
\begin{itemize}
    \item \textbf{LLM-based CTS Architecture}: We leverage LLMs to enable AI-assisted content authoring by generating an easily editable tutoring script from a lesson text, and to automate script orchestration in free-form conversation. The CTS features a learning-by-teaching format with two agents taking on the roles of a student (Ruffle) and a professor (Riley). The human learner engages with these agents, teaching Ruffle with support from Riley.
    \item \textbf{Evaluation of Learning Performance/Experience}: We report findings from two online user studies (N = 200) evaluating the effects of our LLM-driven CTS workflow on learning outcomes and user experience, comparing it to two simpler QA chatbots and reading activity.
    \item \textbf{Evaluation of Interaction/Conversation}: We study usage patterns and conversations and assess their relationships to learning outcomes. We further discuss directions for future system refinements and provide various insights related to the design and evaluation of LLM-based learning technologies.
\end{itemize}
\vspace{-6pt}

\begin{figure}[t]
    \centering
    \includegraphics[width=1.0\textwidth]{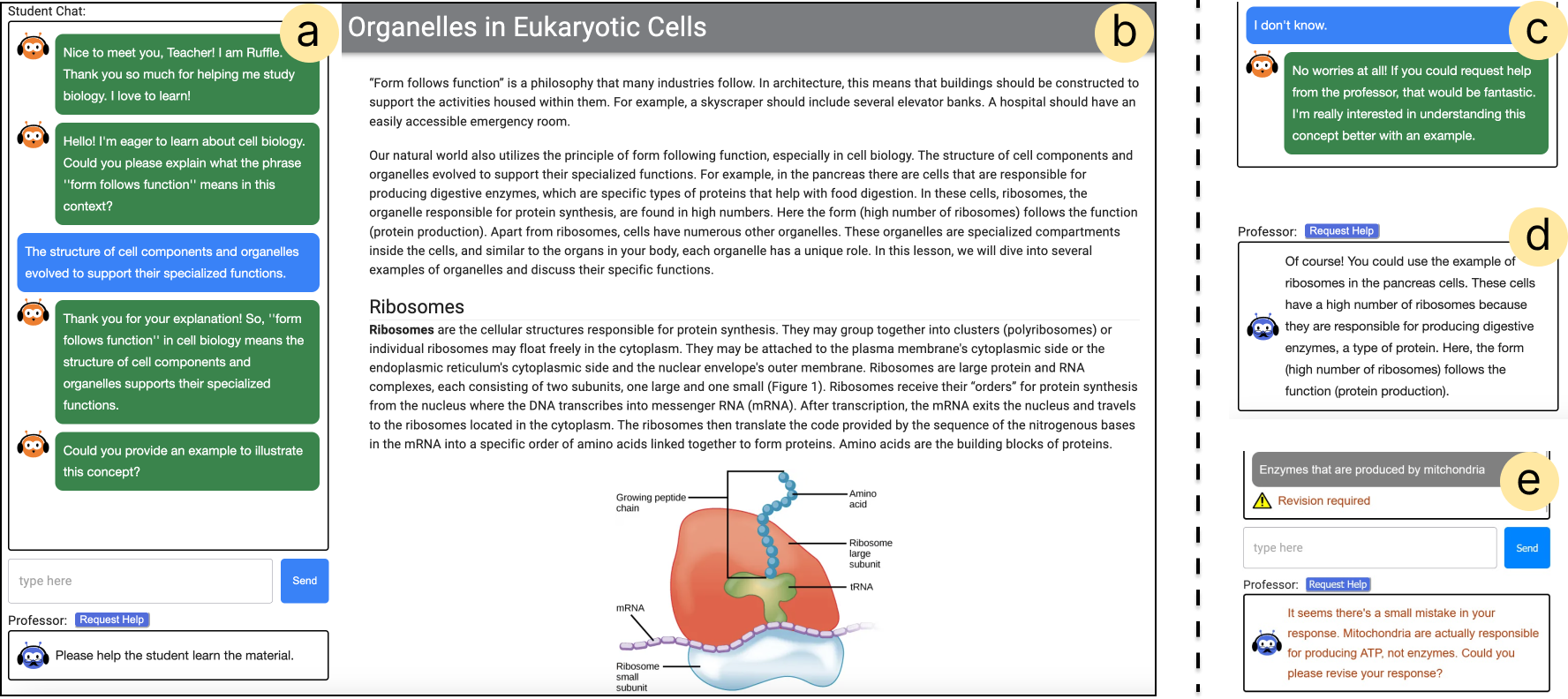}
    \caption{UI of Ruffle\&Riley. (a) Learners are asked to teach Ruffle (student agent) in a free-form conversation and request help as needed from Riley (professor agent). Ruffle tries to guide the learner to articulate the expectations in the tutoring script. (b) The learner can navigate the lesson material during the conversation. (c) Ruffle encourages the learner to explain the content. (d) Riley responds to a help request. (e) Riley detected a misconception and prompts the learner to revise their response.}
    \label{fig:ui}
    \vspace{-10pt}
\end{figure}

\section{Related Work}
\label{sec:related_work}

\paragraph{Conversational Tutoring Systems} Dialog-based learning activities are known to promote high levels of cognitive engagement and to benefit learning outcomes~\cite{Chi2014:Icap}. This motivated the integration of conversational activities into learning technologies. In their systematic review, Paladines and Ramirez~\cite{Paladines2020:Systematic} categorized the design principles underlying existing CTSs into three major categories: (i) expectation misconception tailoring (EMT)~\cite{Nye2014:Autotutor}, (ii) model-tracing (MT)~\cite{Rose2001:Interactive}) and (iii) constraint-based modeling (CBM)~\cite{Mitrovic2005:Effect}. While all three frameworks can benefit learning, they require designers to spend substantial effort configuring the systems for each individual lesson and domain~\cite{Cai2019:Authoring}. Further, due to limitations of underlying NLP techniques, many CTSs struggle to maintain coherent free-form conversations, answer learners' questions, and understand learners' responses reliably~\cite{Nye2014:Autotutor}. In this context, this paper employs recent NLP advances as the foundation for a novel type of LLM-driven CTS to facilitate free-form adaptive dialogues and to alleviate the burdens associated with content authoring.

\paragraph{Content Authoring Tools} One major obstacle to the widespread adoption of ITSs is the complexity of content authoring~\cite{Dermeval2018:Authoring}. For early ITSs, the development ratio (i.e., the number of hours required to create one hour of instructional content) was estimated to vary between 200:1 and 300:1~\cite{Aleven2016:Example}). This motivation the creation of content authoring tools (CATs) to facilitate ITS creation. While a comprehensive survey of CATs is beyond the scope of this paper--for this, we refer to~\cite{Dermeval2018:Authoring}--here we highlighting prior studies that illustrate the ability of existing CATs to reduce authoring times. ASSISTment Builder~\cite{Razzaq2009:Assistment} was developed to support content authoring in a math ITS and enabled a development ratio of 40:1. For model tracing-based ITSs, example tracing~\cite{Aleven2016:Example} has proven itself as an effective technique that depending on the domain enables development ratios between 50:1 and 100:1. In the context of CTSs, multiple CATs have been developed for AutoTutor~\cite{Cai2019:Authoring}, and while we were not able to find concrete development ratio estimates, the authoring of CTS content is still considered to be complex and labor intensive.

Alternative approaches explored the use of learner log data to enhance ITS components~\cite{Barnes2005:Q} as well as machine learning-based techniques for questions and feedback generation~\cite{Kurdi2020:Systematic}. Recent advances in large language models (LLMs) sparked a new wave of research that explores ways in which LLM-based technologies can benefit learners~\cite{Kasneci2023:Chatgpt}. Settings in which LLMs already have been found to be effective include question generation and quality assessment (e.g.,~\cite{Ruan2019:Bookbuddy,Moore2023:Assessing,Jiao2023:Automatic}), feedback generation (e.g.,~\cite{Ruan2019:Bookbuddy,Nguyen2023:Evaluating,Pardos2023:Learning}), answering students' questions (e.g.,~\cite{Lee2023:Dapie}), automated grading (e.g.,~\cite{Botelho2023:Leveraging}), and helping teachers reflect on their teaching (e.g.,~\cite{Markel2023:Gpteach,Demszky2023:Can}). What sets this paper apart from the aforementioned works is that it does not focus on the generation of \textit{individual} ITS components; instead, we propose a system that can automatically induce a \textit{complete ITS workflow}, exhibiting the prototypical inner and outer loop structure~\cite{Vanlehn2006:Behavior}, directly from a lesson text.

\section{System Design and Architecture}
\label{sec:system_overview}

\paragraph{Design Considerations} We approached the design of Ruffle\&Riley with two specific goals in mind: (i) Facilitate an ITS workflow that provides learners with a sequence of questions (outer loop) and meaningful feedback during problem-solving (inner loop); (ii) Streamline the process of configuring the conversational agents for different lesson materials. We reviewed existing CTSs and identified EMT~\cite{Graesser2004:Autotutor} as a suitable design framework. EMT mimics teaching strategies employed by human tutors~\cite{Graesser1995:Collaborative} by associating each question with a list of expectations and anticipated misconceptions. After presenting a question and receiving an initial user response, EMT-based CTSs provide inner loop support (goal (i)) by guiding the conversation via a range of dialogue moves to correct misconceptions and to help the learner articulate the expectations before moving on to the next question (outer loop). While EMT-based CTSs have been shown to be effective in various domains~\cite{Nye2014:Autotutor}, they need to be configured in a labor-intensive process that requires instructional designers to define a \textit{tutoring script} that specifies questions, expectations, misconceptions and other information for each lesson~\cite{Cai2019:Authoring}. For us, tutoring scripts serve as a standardized format for CTS configuration that is easy to read and modify (goal (ii)).

\paragraph{User Interface} An overview of our user interface, together with descriptions of its key elements, is provided by Figure~\ref{fig:ui}. Inspired by the success of learning-by-teaching activities~\cite{duran2017learning,leelawong2008designing}, we decided to orchestrate the conversation in a learning-by-teaching format via two agents taking on the roles of a student (Ruffle) and a professor (Riley). While our design is similar to some CTSs in the AutoTutor family~\cite{Nye2014:Autotutor} that follow a trialogue format, one notable difference is that Riley solely serves as an assistant to the learner by offering assistance and correcting misconceptions. The two agents never talk directly with each other.

\begin{figure}[t]
    \centering
    \includegraphics[width=0.80\textwidth]{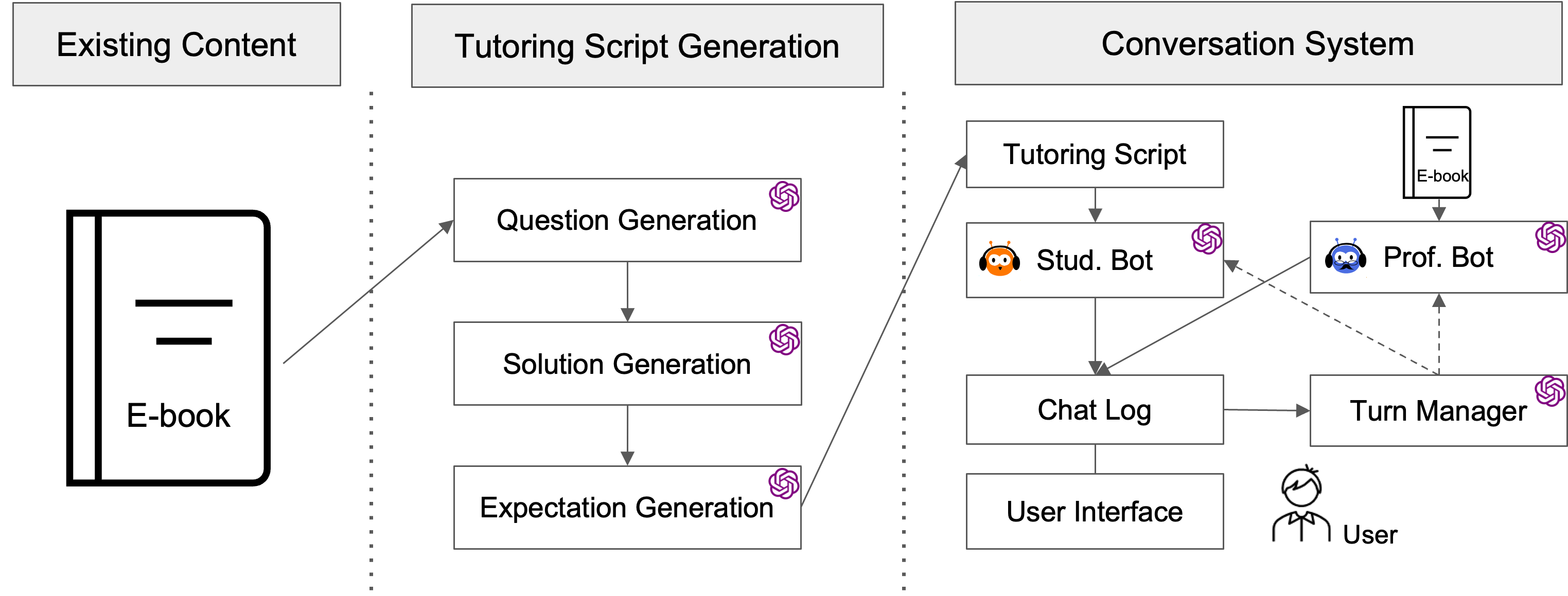}
    \caption{System architecture. Ruffle\&Riley generates a \textit{tutoring script} automatically from a lesson text by executing three separate prompts that induce questions, solutions and expectations for the EMT-based dialog. During the learning process, the script is orchestrated via two LLM-based conversational agents in a free-form dialog that follows the ITS-typical outer and inner loop structure.}
    \label{fig:system}
\end{figure}

\paragraph{AI-Assisted Tutoring Script Authoring} Ruffle\&Riley is capable of generating a tutoring script fully automatically from a lesson text by leveraging GPT4~\cite{Openai2023:Gpt4} (Figure~\ref{fig:system}). This involves a 4-step process: (i) A list of review questions is generated from the lesson text; (ii) For each question, a solution is generated based on question and lesson texts; (iii) For each question, a list of expectations is generated based on question and solution texts; (iv) The final tutoring script is compiled as a list of questions together with related expectations (Figure~\ref{fig:tutoring_script}). The first three steps are implemented via three separate prompts written in a way general enough to support a wide range of lesson materials. The resulting script can be easily modified and revised by instructional designers to meet their needs. Unlike traditional EMT-based CTSs, our tutoring scripts do not attempt to anticipate misconceptions learners might exhibit ahead of time (this is a difficult task even for human domain experts). Instead, we rely on GPT4's to detect and respond to misconceptions in the learner's responses during the active teaching process.

\paragraph{Tutoring Script Orchestration} EMT-based CTSs require the definition of dialog moves and conversational turn management to facilitate coherent conversations, which in itself is a complex authoring process~\cite{Cai2019:Authoring}. Ruffle\&Riley automates the tutoring script orchestration by including descriptions of desirable properties of EMT-based conversations into the agents' prompts and captures the user's state solely via the chat log. The student agent receives the tutoring script as part of its prompt and is instructed to let the user explain the individual questions and to ask follow-ups until all expectations are covered. Ruffle reflects on user responses to show understanding, provides encouragement to the user, and keeps the conversation on topic. In parallel, Riley's prompt contains the lesson text and instructions to offer relevant information after help requests, and to prompt the user to revise their response after detecting incorrect information. Both agents are instructed to keep the conversation positive and encouraging and to not refer to information outside the tutoring script/lesson text. The turn manager coordinates the system's queries to GPT-4. We will open-source our system implementation via GitHub\footnote{GitHub Link} to contribute to ongoing research on effective instructional design of LLM-based learning technologies.

\begin{figure}[t]
    \centering 
    \begin{mdframed}[roundcorner=10, linecolor=black]
    \small{
    \noindent \textbf{Topic 1}: What does the principle ''form follows function'' mean in the context of cell biology? Provide an example to illustrate your answer.

    \vspace{0.8mm}\textit{Fact 1.1}: ''Form follows function'' in cell biology means the structure of cell organelles supports their specialized functions.

    \vspace{0.8mm}\textit{Fact 1.2}: An example is the high number of ribosomes in pancreas cells that produce digestive enzymes, supporting the cell's function of producing proteins.}
    \end{mdframed}
    \caption{\small Tutoring script. To structure the conversational activity, Ruffle\&Riley relies on a pre-generated script featuring a list of questions and related expectations for the EMT-based dialog. Tutoring scripts can be generated automatically from existing lessons text and offer instructional designers a convenient interface for system configuration.}
    \label{fig:tutoring_script}
\end{figure}

\section{Experimental Design}
\label{sec:experimental_design}

We describe the experimental design shared by our two user studies. Studies and participant recruitment were approved by the Institutional Review Board (IRB).

\paragraph{Learning Material} We adapted a Biology lesson on cell organelles from the OpenStax project~\cite{Biology2e:OpenStax}. We decided on this particular lesson because we expected participants to have low prior familiarity with the material to allow for a learning process. The lesson text is accessible to a general audience and covers 640 words.

\paragraph{Conditions} Similar to prior work~\cite{Kopp2012:Improving}, we construct conditions to compare the efficacy of our EMT-based CTS to reading alone and to a QA chatbot with limited dialog. To study potential differences, we equip the QA chatbot with content from different sources under two distinct conditions: one using content designed by a biology teacher and the other using LLM-generated content.

\begin{enumerate}
    \item \textbf{Reading}: Participants study the material without additional support.
    \item \textbf{Teacher QA (TQA)}: Participants study the material and can answer review questions presented by the chatbot. After submitting an answer, participants receive brief feedback about the correctness of their response and a sample solution. Questions and answers were designed by a human teacher.
    \item \textbf{LLM QA (LQA)}: Same as TQA, but questions and answers were generated automatically by the LLM (Section~\ref{sec:system_overview}).
    \item \textbf{Ruffle\&Riley (R\&R)}: Participants study the material while being supported by the two conversational agents (Section~\ref{sec:system_overview}). The system is equipped with the same LLM-generated questions as LQA.
\end{enumerate}

\paragraph{Surveys/Questionnaires} We evaluate system efficacy from two perspectives: \textit{learning performance} and \textit{learning experience}. The \textit{first study} gauges performance via a multiple-choice post-test after the learning session, consisting of five questions written by a biology teacher recruited via Upwork and two questions from OpenStax~\cite{Biology2e:OpenStax}. To evaluate the system more accurately and comprehensively, we conducted the \textit{second study}. In particular, (1) we added a pre-test to test students' prior knowledge and counterbalanced the pre-test and post-test for different students to ensure the difficulty of two tests would not affect the result; (2) we enriched the test questions from only multiple-choice ones to two multiple-choice, three fill-in-the-blank and one free-form response question created to assess participants' deeper understanding of the taught concepts. This revision of the question format was informed by prior work which found the effects of CTSs to be less pronounced in recall-based test formats~\cite{Graesser2004:Autotutor}.

 For both evaluations, learning experience is captured after post-test via a 7-point Likert scale survey that queries participants' perception of engagement, intrusiveness, and helpfulness of the agents, based on prior work~\cite{peng2022crebot}. To ensure data quality, we use two attention checks and one question asking whether participants searched for test answers online. Lastly, we included a demographics questionnaire to understand participants' age, gender, and educational background.

\paragraph{Recruitment} We recruited participants for our user study online via Prolific. Our criteria were: (i) located in the USA; (ii) fluent in English; (iii) possess at least a high-school (HS) degree. Participants were randomly assigned to the learning conditions and were free to drop out at any point in the study. For each of the two evaluation studies, 100 participants completed the task.

\section{Evaluation 1: Initial System Validation}
\label{sec:first_evaluation}

We assess R\&R's ability to facilitate a coherent and structured conversational learning activity. By comparing multiple conditions (Section \ref{sec:experimental_design}), we explore hypotheses related to R\&R's effects on learning performance and experience.

\paragraph{Hypotheses} We explore the following. H1: \textit{Learning Outcomes}: R\&R achieves higher test scores than baseline conditions (H1a); There is no significant difference between TQA and LQA (H1b). H2: \textit{Learning experience}: R\&R achieves higher ratings than baseline conditions in terms of engagement, helpfulness in understanding, remembering, interruption, coherence, support received, and enjoyment (H2a); There are no significant differences between TQA and LQA (H2b).

\paragraph{Participation} As shown in Table~\ref{fig:performance_1}, 30 participants finished the reading condition, 17 finished TQA, 23 finished LQA, and 30 finished R\&R. This imbalance is due to the circular assignment mechanism and dropouts. After filtering participants who failed any of the attention check questions, or who did not rate ``strongly disagree'' when asked whether they looked up test answers, we were left with 58 (male: 33, female: 21, other: 4) out of the 100 participants (15 in reading, 7 in TQA, 15 in LQA, and 21 in R\&R). The age distribution is 18-25 (8), 26-35 (20), 36-45 (18), 46-55 (9), over 55 (3). The degree distribution is: HS or Equiv. (22) Bachelor's/Prof. Degree (25), Master's or Higher (11).

\paragraph{Learning Performance} The post-test consists of seven questions, each worth one point. The mean and standard error in post-test scores for 
each condition are shown by Table~\ref{fig:performance_1}. A one-way ANOVA did not detect significant differences in post-test scores among the four conditions. Therefore, we find support for H1b but not for H1a. Even though not significantly different, we observed that participants in R\&R achieved somewhat higher scores ($5.19 \pm 0.25$) than in TQA ($4.14 \pm 0.83$). We find no significant differences in self-reported prior knowledge.

\begin{table}[t]
    \centering
     \caption{Learning performance across different learning conditions.} \includegraphics[width=1.0\textwidth]{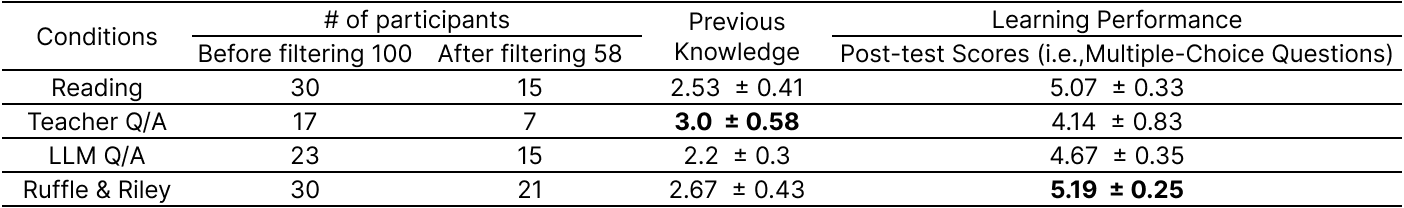}
    \label{fig:performance_1}
\end{table}

\begin{table}[t]
    \centering
    \caption{Learning experience across different conditions. Symbol "*" indicates $p < 0.05$. Symbol "-" indicates that aspect was not asked in the respective condition.}
    \includegraphics[width=1.0\textwidth]{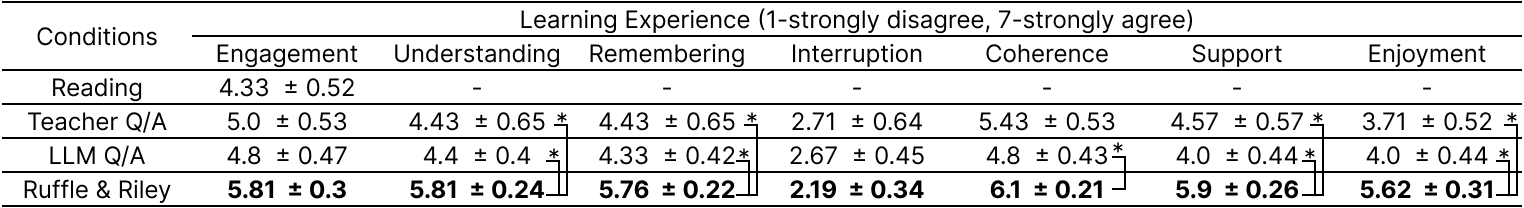}
    \label{fig:experience_1}
\end{table}

\paragraph{Learning Experience} Table~\ref{fig:experience_1} shows participants' learning experience and chatbot interaction ratings. We tested for significance ($p < 0.05$) using one-way ANOVA, followed by Bonferroni post-hoc analysis. We found no significant differences in self-reported engagement levels between the four conditions. However, among the three chatbot conditions, R\&R was rated as significantly more helpful in aiding participants in understanding, remembering the lesson, and providing the support needed to learn. Further, R\&R participants expressed more enjoyment than TQA and LQA participants. In addition, participants found R\&R provided a significantly more coherent conversation than LQA. Interestingly, even though we expected R\&R to be rated as more interrupting, we found no significant differences in perceived interruption among the chatbot conditions. Therefore, H2a is partially supported. In addition, we detected no significant differences in learning experience ratings between LQA and TQA. Thus, we cannot reject H2b.

\paragraph{Insights and Refinements} Reflecting on the results of this first evaluation, we found that R\&R is positively received by its users (Table~\ref{fig:experience_1}). Most importantly, the LLM-based system was able to facilitate coherent free-form conversations across an LLM-generated tutoring script featuring 5 questions and 17 expectations. Even though users were free to end the activity at any point, 17/21 participants completed the entire script. Also, R\&R yields significant learning experience improvements over the more limited QA chatbots (TQA and LQA).

On the other side, R\&R did not lead to significant improvements in learning performance over the reading activity. Further, the mean learning times varied largely between the conditions: reading (4 min), TQA (11 min), LQA (12 min), and R\&R (18 min). Together, this motivated a second evaluation focused on R\&R and reading conditions. For the second evaluation, we performed two revisions: (i) We addressed feedback about the student agent requesting similar information at different points in the conversation by trimming the tutoring script down to 4 questions and 12 expectations; (ii) Because CTSs have been found to be less effective for recall-based test formats~\cite{Nye2014:Autotutor}, we created questions to assess deeper understanding (Section~\ref{sec:experimental_design}).

\section{Evaluation 2: Efficacy and Conversation Analysis}

\begin{table}[t]
    \centering
     \caption{Learning performance for Ruffle\&Riley and reading condition.} \includegraphics[width=1.0\textwidth]{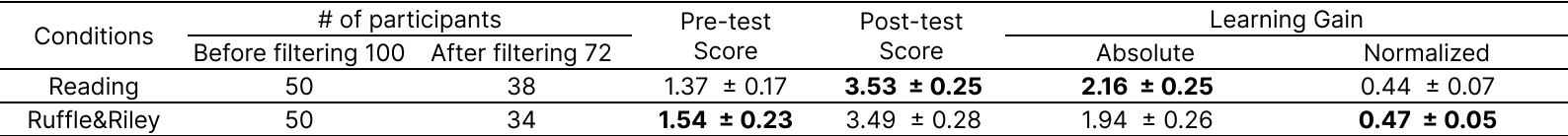}
    \label{fig:performance_2}
\end{table}

\begin{table}[t]
    \centering
    \caption{Learning experience for Ruffle\&Riley and reading condition. The symbol "*" indicates $p < 0.05$.}
    \includegraphics[width=0.9\textwidth]{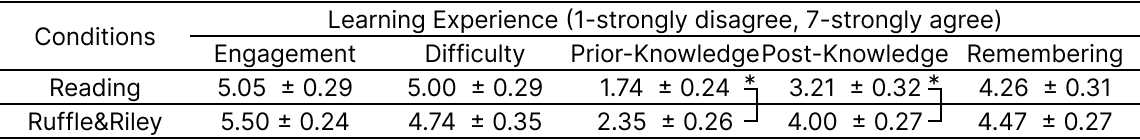}
    \label{fig:experience_2}
\end{table}

In the \textit{second study}, we focus on comparing two conditions: R\&R and reading conditions, which had relatively better learning performance in the \textit{first study} (Table~\ref{fig:performance_1}). We further conduct an in-depth analysis of conversations in R\&R to explore usage patterns and their relations to test performance.

\paragraph{Hypotheses} We explore H1: \textit{Learning Outcomes}: R\&R achieves higher test scores than the reading condition; H2: R\&R achieves better ratings than reading in terms of engagement, understanding, remembering, and perceived difficulty.

\paragraph{Participation} As shown in Table~\ref{fig:performance_2}, reading condition and R\&R condition were each completed by 50 participants. After applying the same filter criteria as in the first evaluation, we were left with 72 (male: 29, female: 43) out of the 100 participants (38 in reading and 34 in R\&R). The age distribution is 18-25 (12), 26-35 (25), 36-45 (15), 46-55 (9), over 55 (11). The degree distribution is HS or Equiv. (23) Bachelor's/Prof. Degree (33), Master's or Higher (16).

\paragraph{Learning Performance} The post-test consists of six questions, each worth one point. The mean and standard error in pre-test and post-test scores as well as derived absolute $(score_{pre} - score_{post})$ and normalized learning gain $(score_{pre} - score_{post}) / (6 - score_{pre})$ measures are provided by Table~\ref{fig:performance_2}. Comparing R\&R and reading condition via one-sided t-tests, we do not detect significant differences ($p < 0.05$) for any of the four measures. Thus, we do not find support for H1.

\paragraph{Learning Experience} Participants' learning experience ratings collected after the post-test are shown in Table~\ref{fig:experience_2}. Different from the tested learning performance, the one-sided t-test unveiled that R\&R users rated their pre- and post-activity knowledge (i.e., perceived knowledge) significantly higher than participants in the reading conditions. While R\&R received advantageous scores in terms of overall engagement, remembering, and task difficulty, these differences could not be established as significant. Together, H2 is partially supported. Engagement scores might not be directly comparable due to large differences in learning times (R\&R (20.8min), reading (5.5min)). Aligning with Evaluation 1, chatbot-specific ratings for R\&R were positive for understanding (5.21), remembering (4.76), interruption (2.41), coherence (6.15), support (5.59), and enjoyment (5.18).

\subsection{Analysing Interaction and Conversations}

\begin{figure}[t]
    \centering
    \includegraphics[width=\textwidth]{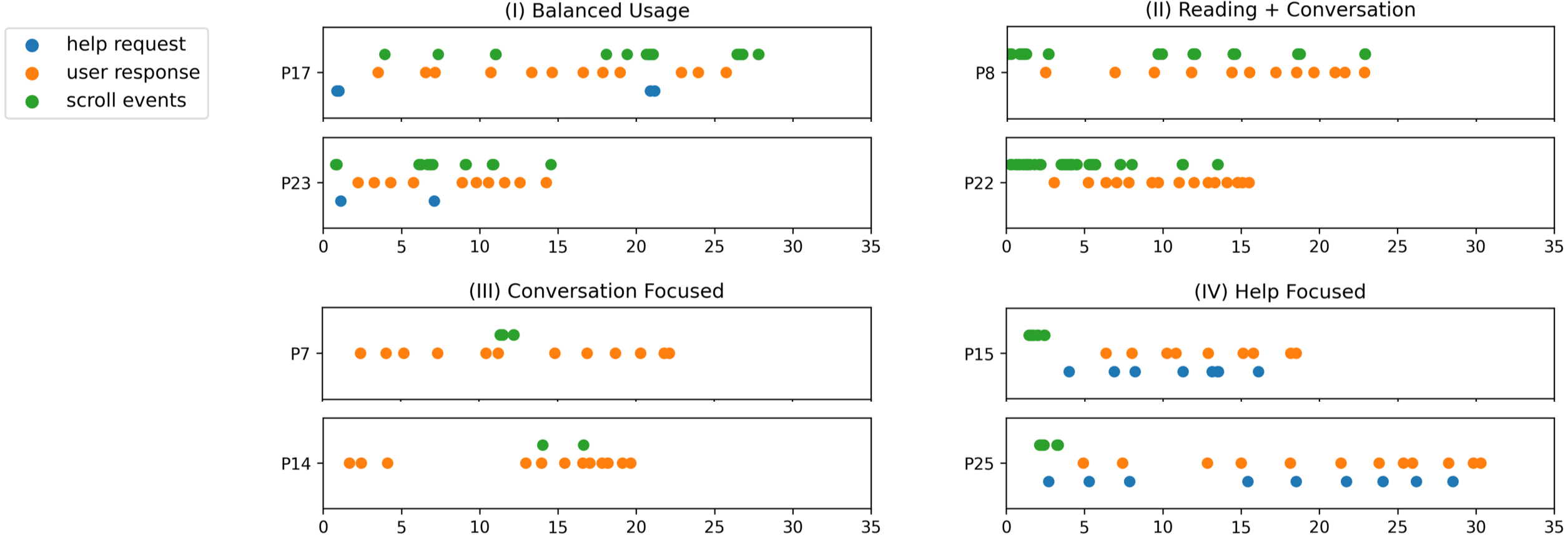}
    \caption{Temporal Interaction Patterns. By visualizing the usage of text navigation, chat response, and help request features over time, we observe four distinct usage patterns.}
    \label{fig:interaction_patterns}
\end{figure}


\begin{table}[t]
\center
\scriptsize
\setlength{\tabcolsep}{4pt}
\caption{Learning performance of Ruffle\&Riley users for each usage patterns.}
\begin{tabular}{cccccc}
\toprule
Usage pattern & Num Users & Pre-Test & Post-Test & Absolute Gain & Relative Gain \\ 
\midrule
Balanced      & 11 & 1.64 $\pm$ 0.48 & 3.27 $\pm$ 0.57 & 1.64 $\pm$ 0.32 & 0.45 $\pm$ 0.10 \\
Read + Conv.  & 13 & 1.62 $\pm$ 0.35 & 3.77 $\pm$ 0.47 & 2.15 $\pm$ 0.37 & 0.53 $\pm$ 0.09 \\
Conv. Focused &  4 & 0.62 $\pm$ 0.24 & 4.12 $\pm$ 0.52 & 3.50 $\pm$ 0.46 & 0.65 $\pm$ 0.09 \\
Help Focused  &  3 & 0.67 $\pm$ 0.33 & 2.17 $\pm$ 0.44 & 1.50 $\pm$ 0.50 & 0.28 $\pm$ 0.09 \\
\bottomrule
\end{tabular}
\label{tab:pattern_performance}
\end{table}
\vspace{-5pt}

\paragraph{Interactions} We analyze interaction log data in R\&R. First, 31/34 participants completed the full conversational workflow. Users submitted on average $1.71 \pm 0.45$ help requests and received $1.77 \pm 0.23$ response revision requests from Riley. Second, by evaluating temporal usage of the conversation, scrolling, and help request features, we observe four distinct system usage patterns among the 31 participants that completed the activity (Figure~\ref{fig:interaction_patterns}): (I) balanced feature usage; (II) conversation and reading only; (III) focus on conversation; (IV) focus on requesting help. Table~\ref{tab:pattern_performance} provides learning performance measures for each group. While sample sizes are too small to draw conclusions, we observe that the conversation-focused users (group (III)) achieve the highest learning gains. Further, the non-help seeking users (group (II) and (III)) achieve higher performance measures than the help-seeking users (group (I) and (IV)).

\paragraph{Conversation} We analyze conversation log data in R\&R. First, we evaluate the correct execution of the EMT-based dialogues. While all 31 participants went through all four questions in the tutoring script, we noticed that seven conversations omitted one to two expectations. Further, in nine conversations, the student agent requested similar information at different points in the session, often when users wrote long responses covering multiple expectations at once. Another issue uncovered was that the system was often lenient towards user responses that only covered parts of one expectation (e.g., mention cellular respiration but do not explain its in- and outputs). An evaluation of all conversations verified the factual correctness of the GPT-4-based agents' responses. While we find that R\&R facilitates coherent free-form conversational tutoring, future revisions are needed to enhance the system's ability to provide users with effective feedback.  

Second, we assess correlations between conversation features and participants' learning performance (Table~\ref{tab:correlation_analysis}). \textit{Learning time} and the number of \textit{words} in user explanations both show positive correlations to the performance measures. The number of submitted \textit{help} and received \textit{revision} requests exhibit negative correlations. We observe no significant correlations between number of user \textit{responses} and performance measures. In summary, the above analysis indicates that the way participants' engage with the system affects their learning outcomes.


\begin{table}[t]
\center
\scriptsize
\setlength{\tabcolsep}{4pt}
\caption{Pearson correlation analysis between conversation features and performance.}
\begin{tabular}{lrrrr}
\toprule
Feeature & Pre-Test & Post-Test & Absolute Gain & Relative Gain \\ 
\midrule
\# User Messages & -0.06 (p = 0.75) &  0.12 (p = 0.54) &  0.22 (p = 0.24) &  0.11 (p = 0.54) \\
\# Help Requests & -0.14 (p = 0.44) & -0.35 (p = 0.05) & -0.32 (p = 0.08) & -0.34 (p = 0.06) \\
\# Revisions     & -0.35 (p < 0.05) & -0.20 (p = 0.27) &  0.10 (p = 0.58) & -0.13 (p = 0.50) \\
\# Words         &  0.24 (p = 0.19) &  0.38 (p = 0.04) &  0.26 (p = 0.17) &  0.40 (p = 0.02) \\
Learning Time (min)       & -0.17 (p = 0.34) &  0.16 (p = 0.35) &  0.39 (p = 0.02) &  0.26 (p = 0.13) \\
\bottomrule
\end{tabular}
\label{tab:correlation_analysis}
\end{table}

\section{Discussion and Future Work}
\label{sec:discussion}

Ruffle\&Riley is a conversational tutoring system (CTS) that leverages recent advances in large language models (LLMs) to generate tutoring scripts automatically using existing lessons texts. These tutoring scripts define conversational learning activities which are orchestrated via two LLM-based agents (Ruffle\&Riley) acting as a student and a professor. The human learner engages with the system by explaining a series of topics to the student (Ruffle) while being supported by the professor (Riley). Our user studies verified the system's ability to facilitate coherent free-form conversational tutoring. This highlights the potential of generative-AI-assisted content authoring for lowering resource requirements of CTS content development~\cite{Cai2019:Authoring}, promoting the design of learning activities meeting the needs of a wider diversity of learners and a broader range of subjects.

Our \textit{first} study (N = 100), evaluates Ruffle\&Riley's ability to support biology lessons comparing the system to QA chatbots offering limited feedback and a reading activity. In terms of learning experience, Ruffle\&Riley users reported significantly higher ratings in terms of understanding, remembering, helpfulness of support and enjoyment.
Still, corroborating prior research, the recall-focused multiple-choice post-test did not detect significant differences in learning outcomes between conversational tutoring and reading~\cite{Graesser2004:Autotutor}. 
A \textit{second} study (N = 100) compared Ruffle\&Riley and reading condition using questions created to assess deeper understanding. Again. we detected no significant differences in learning outcomes between the conditions even though Ruffle\&Riley users required on average more time (20.8 min) than participants in reading condition (5.5 min). 

We performed an in-depth evaluation of interaction and conversation log data to better understand how usage of Ruffle\&Riley relates to learning outcomes. By studying temporal usage of conversation, scrolling and help request features we were able to identify four distinct system usage patterns. Interestingly, we found that the users that focused on conversation and that did not request help achieved the highest learning gains indicating a Doer effect~\cite{Koedinger2015:Learning}. The worst performing group, exhibited gaming behavior requesting help before each response. We further identified positive correlations between learning outcomes and number of \textit{words} in users' explanations and overall learning \textit{time}.
Future work will explore revisions to nudge users towards active practice and to mitigate gaming behavior~\cite{Baker2006:Adapting}.

Reviewing the conversational logs, we found Ruffle\&Riley to be receptive for partial explanations that miss important information (e.g., mention cellular respiration without explaining its in- and outputs) moving the conversation ahead too quickly. Future work will focus on enhancing the system's ability to provide feedback to help users elicit all information. While hallucination and biased outputs are well recorded problems for LLM-based learning technologies~\cite{Kasneci2023:Chatgpt}, we highlight \textit{affirmation of imprecise user responses} as additional challenge. 

The present study is subject to several limitations: First, the system was evaluated in an online user study conducted via Prolific with adult participants exhibiting diverse demographics (i.e., age and education). Current findings focus on a broad population of online users and might not generalize to more specific populations (e.g., K12 or college students). Still, this environment enabled us to identify limitations of our system emphasizing the need for future research on instructional design principles~\cite{Koedinger2013:Instructional} for LLM-based CTSs to improve effects on learning performance and efficiency~\cite{Kopp2012:Improving}.
Second, before evaluating Ruffle\&Riley with younger learners, we need to certify safe and trustworthy system behavior adding to the system's existing mechanisms designed to ensure factual correctness of information that surfaces during conversations~\cite{Kasneci2023:Chatgpt}. 
Third, participants only took part in a single learning session and might require time to adapt to the workflow.
Lastly, while generative AI-based learning technologies like our GPT-4 based system show promise, they also incur regular costs due to API calls, raising important questions about equity and accessibility in educational contexts~\cite{Kasneci2023:Chatgpt}.

\section{Conclusion}
\label{sec:conclusion}

In this paper we introduced Ruffle\&Riley, a novel type of LLM-based CTS, enabling AI-assisted content authoring and free-form conversational tutoring. We adopted Expectation Misconception Tailoring (EMT)~\cite{Nye2014:Autotutor} as design framework to facilitate structured conversational learning via pre-generated tutoring scripts. Importantly, scripts can be generated from existing lesson texts and can be revised by instructors based on their needs. We conducted two online users studies (N = 200) verifying Ruffle\&Riley's ability to host conversational tutoring and to promote a positive learning experience. Still, our studies did not reveal significant differences in learning outcomes compared to the shorter reading activity, highlighting the importance of systematic evaluations of generative-AI-based technologies. An analysis of interaction log data motivates future system refinements to enhance Ruffle\&Riley's ability to provide targeted feedback in response to imprecise user explanations. Our system architecture and experiments provide various insights for the design and evaluation of future CTSs.

%
%
%
\bibliographystyle{splncs04}
\bibliography{bibliography}

\begin{thebibliography}{10}
\providecommand{\url}[1]{\texttt{#1}}
\providecommand{\urlprefix}{URL }
\providecommand{\doi}[1]{https://doi.org/#1}

\bibitem{Aleven2016:Example}
Aleven, V., McLaren, B.M., Sewall, J., Van~Velsen, M., Popescu, O., Demi, S.,
  Ringenberg, M., Koedinger, K.R.: Example-tracing tutors: Intelligent tutor
  development for non-programmers. Int. Journal of Artificial Intelligence in
  Education  \textbf{26},  224--269 (2016)

\bibitem{Baker2006:Adapting}
Baker, R.S.d., Corbett, A.T., Koedinger, K.R., Evenson, S., Roll, I., Wagner,
  A.Z., Naim, M., Raspat, J., Baker, D.J., Beck, J.E.: Adapting to when
  students game an intelligent tutoring system. In: Intelligent Tutoring
  Systems: 8th International Conference, ITS 2006, Jhongli, Taiwan, June 26-30,
  2006. Proceedings 8. pp. 392--401. Springer (2006)

\bibitem{Barnes2005:Q}
Barnes, T.: The q-matrix method: Mining student response data for knowledge.
  In: American association for artificial intelligence 2005 educational data
  mining workshop. pp.~1--8. AAAI Press, Pittsburgh, PA, USA (2005)

\bibitem{Botelho2023:Leveraging}
Botelho, A., Baral, S., Erickson, J.A., Benachamardi, P., Heffernan, N.T.:
  Leveraging natural language processing to support automated assessment and
  feedback for student open responses in mathematics. Journal of Computer
  Assisted Learning  (2023)

\bibitem{Cai2019:Authoring}
Cai, Z., Hu, X., Graesser, A.C.: Authoring conversational intelligent tutoring
  systems. In: Adaptive Instructional Systems: First Int. Conf., AIS 2019,
  Orlando, FL, USA. pp. 593--603. Springer (2019)

\bibitem{Chi2014:Icap}
Chi, M.T., Wylie, R.: The icap framework: Linking cognitive engagement to
  active learning outcomes. Educational psychologist  \textbf{49}(4),  219--243
  (2014)

\bibitem{Biology2e:OpenStax}
Clark, M.A., Douglas, M., Choi, J.: Biology 2e. OpenStax (2018)

\bibitem{Demszky2023:Can}
Demszky, D., Liu, J., Hill, H.C., Jurafsky, D., Piech, C.: Can automated
  feedback improve teachers’ uptake of student ideas? evidence from a
  randomized controlled trial in a large-scale online course. Educational Eval.
  and Policy Analysis  (2023)

\bibitem{Dermeval2018:Authoring}
Dermeval, D., Paiva, R., Bittencourt, I.I., Vassileva, J., Borges, D.:
  Authoring tools for designing intelligent tutoring systems: a systematic
  review of the literature. Int. Journal of Artificial Intelligence in
  Education  \textbf{28},  336--384 (2018)

\bibitem{duran2017learning}
Duran, D.: Learning-by-teaching. evidence and implications as a pedagogical
  mechanism. Innovations in Education and Teaching International
  \textbf{54}(5),  476--484 (2017)

\bibitem{Graesser2004:Autotutor}
Graesser, A.C., Lu, S., Jackson, G.T., Mitchell, H.H., Ventura, M., Olney, A.,
  Louwerse, M.M.: Autotutor: A tutor with dialogue in natural language.
  Behavior Research Methods, Instruments, \& Computers  \textbf{36},  180--192
  (2004)

\bibitem{Graesser1995:Collaborative}
Graesser, A.C., Person, N.K., Magliano, J.P.: Collaborative dialogue patterns
  in naturalistic one-to-one tutoring. Applied cognitive psychology
  \textbf{9}(6),  495--522 (1995)

\bibitem{Huang2016:Intelligent}
Huang, X., Craig, S.D., Xie, J., Graesser, A., Hu, X.: Intelligent tutoring
  systems work as a math gap reducer in 6th grade after-school program.
  Learning and Individual Differences  \textbf{47},  258--265 (2016)

\bibitem{Jiao2023:Automatic}
Jiao, Y., Shridhar, K., Cui, P., Zhou, W., Sachan, M.: Automatic educational
  question generation with difficulty level controls. In: Int. Conf. on
  Artificial Intelligence in Education. pp. 476--488. Springer (2023)

\bibitem{Kasneci2023:Chatgpt}
Kasneci, E., Se{\ss}ler, K., K{\"u}chemann, S., Bannert, M., Dementieva, D.,
  Fischer, F., Gasser, U., Groh, G., G{\"u}nnemann, S., H{\"u}llermeier, E.,
  et~al.: Chatgpt for good? on opportunities and challenges of large language
  models for education. Learning and individual differences  \textbf{103},
  102274 (2023)

\bibitem{Koedinger2013:Instructional}
Koedinger, K.R., Booth, J.L., Klahr, D.: Instructional complexity and the
  science to constrain it. Science  \textbf{342}(6161),  935--937 (2013)

\bibitem{Koedinger2015:Learning}
Koedinger, K.R., Kim, J., Jia, J.Z., McLaughlin, E.A., Bier, N.L.: Learning is
  not a spectator sport: Doing is better than watching for learning from a
  mooc. In: Proceedings of the second (2015) ACM conference on learning@ scale.
  pp. 111--120 (2015)

\bibitem{Kopp2012:Improving}
Kopp, K.J., Britt, M.A., Millis, K., Graesser, A.C.: Improving the efficiency
  of dialogue in tutoring. Learning and Instruction  \textbf{22}(5),  320--330
  (2012)

\bibitem{Kulik2016:Effectiveness}
Kulik, J.A., Fletcher, J.D.: Effectiveness of intelligent tutoring systems: A
  meta-analytic review. Review of Educational Research  \textbf{86}(1),  42--78
  (2016)

\bibitem{Kurdi2020:Systematic}
Kurdi, G., Leo, J., Parsia, B., Sattler, U., Al-Emari, S.: A systematic review
  of automatic question generation for educational purposes. Int. Journal of
  Artificial Intelligence in Education  \textbf{30},  121--204 (2020)

\bibitem{Lee2023:Dapie}
Lee, Y., Kim, T.S., Kim, S., Yun, Y., Kim, J.: Dapie: Interactive step-by-step
  explanatory dialogues to answer children’s why and how questions. In:
  Proceedings of the 2023 CHI Conf. on Human Factors in Computing Systems. pp.
  1--22 (2023)

\bibitem{leelawong2008designing}
Leelawong, K., Biswas, G.: Designing learning by teaching agents: The betty's
  brain system. Int. Journal of Artificial Intelligence in Education
  \textbf{18}(3),  181--208 (2008)

\bibitem{Markel2023:Gpteach}
Markel, J.M., Opferman, S.G., Landay, J.A., Piech, C.: Gpteach: Interactive ta
  training with gpt-based students. In: Proceedings of the 10th ACM Conf. on
  Learning@Scale. p. 226–236. ACM, New York, NY, USA (2023)

\bibitem{Mitrovic2005:Effect}
Mitrovic, A.: The effect of explaining on learning: a case study with a data
  normalization tutor. In: AIED. pp. 499--506 (2005)

\bibitem{Moore2023:Assessing}
Moore, S., Nguyen, H.A., Chen, T., Stamper, J.: Assessing the quality of
  multiple-choice questions using gpt-4 and rule-based methods. In: European
  Conference on Technology Enhanced Learning. pp. 229--245. Springer (2023)

\bibitem{Nguyen2023:Evaluating}
Nguyen, H.A., Stec, H., Hou, X., Di, S., McLaren, B.M.: Evaluating chatgpt’s
  decimal skills and feedback generation in a digital learning game. In:
  European Conference on Technology Enhanced Learning. pp. 278--293. Springer
  (2023)

\bibitem{Nye2014:Autotutor}
Nye, B.D., Graesser, A.C., Hu, X.: Autotutor and family: A review of 17 years
  of natural language tutoring. Int. Journal of Artificial Intelligence in
  Education  \textbf{24},  427--469 (2014)

\bibitem{Openai2023:Gpt4}
OpenAI: Gpt-4 technical report (2023)

\bibitem{Paladines2020:Systematic}
Paladines, J., Ramirez, J.: A systematic literature review of intelligent
  tutoring systems with dialogue in natural language. IEEE Access  \textbf{8},
  164246--164267 (2020)

\bibitem{Pardos2023:Learning}
Pardos, Z.A., Bhandari, S.: Learning gain differences between chatgpt and human
  tutor generated algebra hints. arXiv preprint arXiv:2302.06871  (2023)

\bibitem{peng2022crebot}
Peng, Z., Liu, Y., Zhou, H., Xu, Z., Ma, X.: Crebot: Exploring interactive
  question prompts for critical paper reading. Int. Journal of Human-Computer
  Studies  \textbf{167} (2022)

\bibitem{Razzaq2009:Assistment}
Razzaq, L., Patvarczki, J., Almeida, S.F., Vartak, M., Feng, M., Heffernan,
  N.T., Koedinger, K.R.: The assistment builder: Supporting the life cycle of
  tutoring system content creation. IEEE Transactions on Learning Technologies
  \textbf{2}(2),  157--166 (2009)

\bibitem{Rose2001:Interactive}
Ros{\'e}, C.P.: Interactive conceptual tutoring in atlas-andes. Artif. Intell.
  in Education: AI-Ed in the Wired and Wireless Future pp. 256--266 (2001)

\bibitem{Ruan2019:Bookbuddy}
Ruan, S., Willis, A., Xu, Q., Davis, G.M., Jiang, L., Brunskill, E., Landay,
  J.A.: Bookbuddy: Turning digital materials into interactive foreign language
  lessons through a voice chatbot. In: Proceedings of the sixth (2019) ACM
  conf. on learning@ scale. pp.~1--4 (2019)

\bibitem{Vanlehn2006:Behavior}
VanLehn, K.: The behavior of tutoring systems. Int. journal of artificial
  intelligence in education  \textbf{16}(3),  227--265 (2006)

\end{thebibliography}

\end{document}